\documentclass[letterpaper, 10pt]{ieeeconf}
\IEEEoverridecommandlockouts

\usepackage{cite}
\usepackage{amsmath,amssymb,amsfonts}
\usepackage{bm}
\usepackage{algorithmic}
\usepackage{graphicx}
\usepackage{textcomp}
\usepackage{xcolor,color}
\usepackage[ruled,vlined]{algorithm2e}
\usepackage[top=0.833in, left=0.667in, right=0.667in, bottom=0.597in, includefoot]{geometry}
\usepackage{hyperref}
\usepackage{tikz}
\usepackage{gensymb}
\usepackage{arydshln}
\usetikzlibrary{positioning}

\usepackage{amsthm}
    
\makeatletter
\let\MYcaption\@makecaption
\makeatother 
\usepackage[font=footnotesize]{subcaption}
\makeatletter
\let\@makecaption\MYcaption
\makeatother

\newcommand\undermat[2]{%
  \makebox[0pt][l]{$\smash{\underbrace{\phantom{%
    \begin{matrix}#2\end{matrix}}}_{\text{$#1$}}}$}#2}
\usepackage{balance}

\newtheorem{remark}{Remark}
\newtheorem{problem}{Problem}

\newcommand{\hl}[1]{#1}

\newcommand{\q}{\bm{q}}
\newcommand{\qd}{\dot{\q}}
\newcommand{\qdd}{\ddot{\q}}
\newcommand{\btau}{\bm{\tau}}

\newcommand{\M}{\bm{M}}
\newcommand{\C}{\bm{C}}

\newcommand{\Q}{\bm{Q}}
\newcommand{\R}{\bm{R}}

\newcommand{\x}{\bm{x}}
\newcommand{\xd}{\dot{\x}}

\renewcommand{\u}{\bm{u}}

\newcommand{\p}{\bm{p}}
\newcommand{\g}{\bm{g}}

\newcommand{\W}{\bm{W}}
\newcommand{\K}{\bm{K}}

\begin{document}

\title{Mini Cheetah, the Falling Cat: A Case Study in Machine Learning \\ and Trajectory Optimization for Robot Acrobatics}

\author{Vince Kurtz, He Li, Patrick M. Wensing, and Hai Lin\\[-5ex]~
\thanks{The authors are with the Departments of Electrical Engineering (Kurtz and Lin) and Aerospace and Mechanical Engineering (Li and Wensing), University of Notre Dame, Notre Dame, IN, 46556 USA. \texttt{\{vkurtz,hli25,pwensing,hlin1\}@nd.edu}}
\thanks{This work was supported by NSF Grants IIS-1724070, CNS-1830335, IIS2007949, CMMI-1835186.}
}

\maketitle
\thispagestyle{empty}

\begin{abstract}
    Seemingly in defiance of basic physics, cats consistently land on their feet after falling. In this paper, we design a controller that lands the Mini Cheetah quadruped robot on its feet as well. Specifically, we explore how trajectory optimization and machine learning can work together to enable highly dynamic bioinspired behaviors. We find that a reflex approach, in which a neural network learns entire state trajectories, outperforms a policy approach, in which a neural network learns a mapping from states to control inputs. We validate our proposed controller in both simulation and hardware experiments, and are able to land the robot on its feet from falls with initial pitch angles between -90 and 90 degrees.
\end{abstract}

\section{Introduction and Related Work}\label{sec:intro}

The ability of animals like cats and squirrels to consistently land on their feet after falling has long fascinated scientists and engineers \cite{kane1969dynamical,montgomery1993gauge,batterman2003falling,enos1993dynamics,frohlich1980physics}. This seemingly physics-defying feat, in which the animal starts and ends with zero angular velocity but somehow completely changes orientation, relies on exploiting conservation of angular momentum, a non-holonomic constraint. Specifically, cats counter-rotate their front and back halves, alternately tucking and extending their legs to change orientation while maintaining constant angular momentum \cite{kane1969dynamical}.

In this paper, we explore whether similar rapid re-orientation maneuvers are possible for torque-controlled quadruped robots like the Mini Cheetah \cite{katz2019mini}. There are several motivations for this exploration. First, we are interested in applying the underlying physical principles used by animals to control systems with different morphologies (our quadruped robot doesn't have a flexible spine, so directly mimicking a cat isn't an option). Second, we are interested in testing the limits of our hardware and controller design methods. Specifically, we are interested in comparing different ways to combine trajectory optimization and machine learning for aggressive control of a high-dimensional nonlinear system. Finally, this is closely related to practical problems such as reorientation in zero gravity \cite{rudin2021cat}.

\begin{figure}
    \centering
    \begin{subfigure}{0.45\linewidth}
        \centering
        \includegraphics[width=0.65\linewidth]{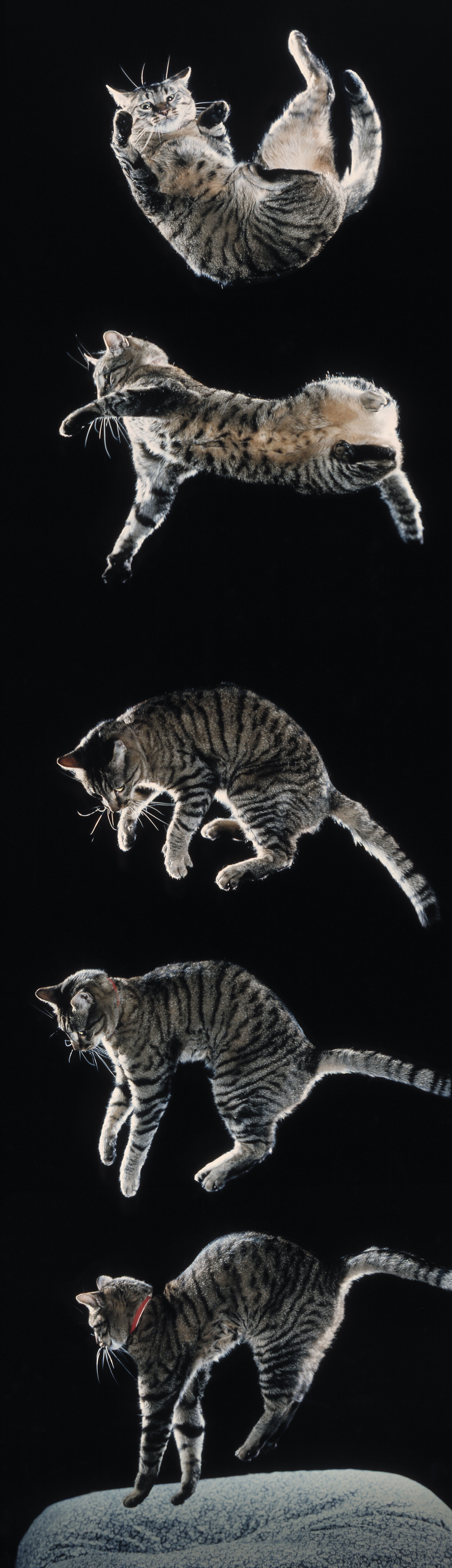}
        \caption{A cat \cite{catpicture}}
        \label{fig:cat_composite}
    \end{subfigure}
    \begin{subfigure}{0.45\linewidth}
        \centering
        \includegraphics[width=0.65\linewidth]{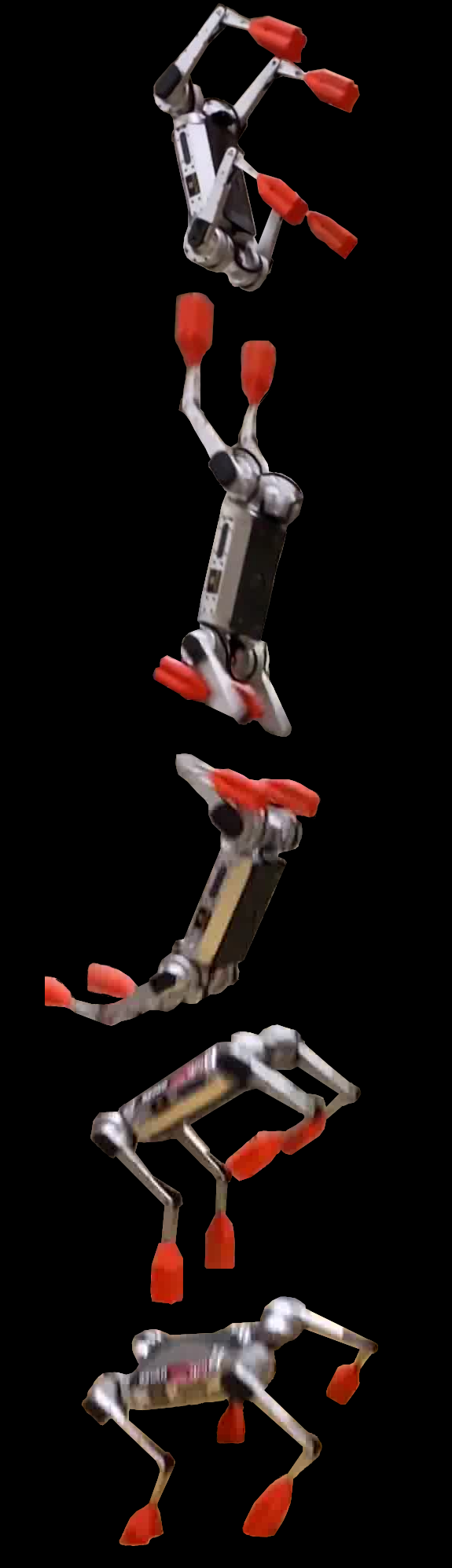}
        \caption{Mini Cheetah}
        \label{fig:minicheetah_composite}
    \end{subfigure}
    \caption{Cats have an amazing ability to rapidly reorient themselves in midair, allowing them to land on their feet after falling. In this paper, we use a combination of trajectory optimization and supervised learning to achieve similar mid-air reorientation for a Mini Cheetah quadruped robot. }
    \label{fig:composite_picture}
\end{figure}

Much prior work on cat-like landings for robots has focused on biomimetic strategies, where a robot is specifically designed to land on its feet, often using a flexible back to imitate cats \cite{shields2013falling,bingham2014orienting}. \hl{Other work has compared inertial reorientation using flywheels, tails, and/or limbs \cite{libby2016comparative}. Their respective benefits and drawbacks were studied with RHex \cite{libby2016comparative}, a hexapod with single-DoF recirculating legs. Cats themselves do not primarily rely on their tails to land on their feet \cite{kane1969dynamical}, instead relying on the articulation of limbs and body for mid-flight reorientation. In a similar manner, the goal of this work is to land a general-purpose quadruped robot (Mini Cheetah) on its feet through inertia shaping via its articulated limbs.} 

With this in mind, our project is closest related to the excellent work of Rudin et al. \cite{rudin2021cat}, which develops a Reinforcement Learning (RL) controller for a quadruped robot in low-gravity, and shows how this controller can be used for cat-like reorientation. The primary difference between this work and \cite{rudin2021cat} is the fact that we are constrained to work in earth's gravitational field, which requires more dramatic motions at the edge of the robot's control authority and joint limits. This motivates us to use a different set of computational tools; specifically, trajectory optimization and supervised learning. 

Trajectory optimization is widely used for planning and control of high degree-of-freedom robots like quadrupeds and humanoids. Standard trajectory optimization methods are especially well-suited to systems with smooth dynamics: in this regard the falling cat problem is ideal, as mid-air reorientation does not involve non-smooth contact dynamics. Specifically, given an initial state of the robot, we can use Differential Dynamic Programming (DDP) \cite{mayne1966second,tassa2012synthesis,li2020hybrid} to find a trajectory that lands the robot on its feet. Unfortunately, however, trajectory optimization can be slow, especially when a good initial guess is not available. In the context of conventional locomotion strategies like walking and running, such initial guesses can be generated by simplified models \cite{wieber2016modeling}, but such templates are not readily available for tasks that require whole-body shape changes like in mid-air acrobatics. 

We overcome this limitation by combining trajectory optimization with machine learning. There has been considerable research on this topic in recent years. One approach is to focus on the cost function, regularizing the running cost with learned heuristics \cite{bledt2017policy} or using a neural network to set the terminal cost \cite{lowrey2018plan}. Others combine trajectory optimization with RL, using trajectory optimization for low-level control \cite{xie2021glide} or to rapidly explore the state space \cite{lowrey2018plan}. Closest to our approach are learned initializations for trajectory optimization \cite{mansard2018using}.

In our case, we solve the trajectory optimization problem offline for many initial conditions and use a neural network to ``memorize'' the results. At runtime, this network outputs a trajectory that lands the robot on its feet. Specifically, we investigate two methods of storing offline optimization results: as a \textit{policy}, which maps states to control torques; and as a \textit{reflex}, which maps the initial state to a trajectory that is then tracked with a PD+ controller. We find that the reflex method requires less training data and outperforms the policy method. 

\hl{This work contributes the first example of landing a general-purpose quadruped robot on its feet like a cat (in real-time under earth's gravity)\footnote{Video summary: \url{https://youtu.be/U5sQSqudWZA}}.
In doing so, we carry out a case-study in using trajectory optimization for bioinspired controller design: while we don't directly imitate a cat's motion (Fig.~\ref{fig:cat_composite}), our controller exploits the same underlying physical mechanism---inertia shaping under conservation of angular momentum---to accomplish the task. This demonstrates the capabilities of the Mini Cheetah hardware platform on a demanding acrobatic task. Specifically, we show that Mini Cheetah is capable of mid-air rotations of up to $90\degree$ in 0.5s, though performing such maneuvers requires adding weights to the feet and operating at the torque limits of the robot.}

The remainder of this paper is organized as follows: we provide a formal problem statement in Section~\ref{sec:problem_formulation}; our proposed control strategy is described in Section~\ref{sec:methodology}; simulation experiments, including a comparison of policy and reflex methods, are presented in Section~\ref{sec:simulation}; and hardware experiments are presented in Section~\ref{sec:hardware}. We provide a detailed discussion of the advantages and limitations of our proposed control scheme along with areas for potential future research in Section~\ref{sec:discussion}, and conclude with Section~\ref{sec:conclusion}. 

\section{Problem Formulation}\label{sec:problem_formulation}

Our goal is to land Mini Cheetah on its feet after falling. Specifically, we focus on rotations in the sagittal plane and initial orientations (pitch angles) between $-90\degree$ (nose facing up) and $90\degree$ (nose down). We assume the robot is dropped from roughly 1.5m above the ground, meaning the recovery motion needs to be completed in about 0.5s. 

We know that in order to change orientation via conservation of angular momentum, the robot will need to vary its distribution of mass as it falls. Unfortunately, the legs of Mini Cheetah were specifically designed to be very lightweight \cite{katz2019mini}, so moving them around doesn't change the distribution of mass very much. To avoid this issue, we equip the robot with special heavy ``boots''. These boots are 3D printed and each hold two rolls of standard American nickels, leading to a total additional mass of roughly 500g per foot. 

We assume that the robot is constrained to move in the sagittal plane, and that the front and back pairs of legs move in unison. The dynamics of the robot in free-fall (with heavy boots) can then be written in standard ``manipulator'' form as
\begin{equation}\label{eq:dynamics}
    \M
    \begin{bmatrix}
        \ddot{\p} \\
        \ddot{\theta} \\
        \qdd
    \end{bmatrix} + 
    \C
    \begin{bmatrix}
        \dot{\p} \\
        \dot{\theta} \\
        \qd
    \end{bmatrix} + 
    \g = 
    \begin{bmatrix}
        \bm{0} \\
        0  \\
        \btau
    \end{bmatrix}
\end{equation}
where $\p \in \mathbb{R}^2$ is the position of the center-of-mass, $\theta \in \mathbb{R}$ is the orientation (pitch) of the body frame, $\q \in \mathbb{R}^4$ are joint angles, and $\btau \in \mathbb{R}^4$ are applied joint torques (control inputs). $\M$ is the positive-definite inertia matrix, $\C$ is the Coriolis matrix, and $\g$ characterizes the influence of gravity.  

Since the position $\p$ is not important for the reorientation task in free-fall, we ignore it in most of our computations and consider the robot's state to be given by $\x = [\theta, \dot{\theta}, \q^T, \qd^T]^T$. We can then write the dynamics in standard form as
\begin{equation}\label{eq:dynamics_general}
    \xd = f(\x,\u),
\end{equation}
where $\u = \btau$ are control inputs and $f$ is a smooth map. 

We denote the state at a given timestep $t$ as $\x(t)$, and the entire state trajectory as $\x(\cdot)$. Numerically, we represent such trajectories simply with samples at 500Hz, though other options such as splines could also be used \cite{kelly2017introduction}. 

We can now present the problem more formally as follows:

\begin{problem}[Falling Cat]\label{prob:falling_cat}
    Given an initial state $\x(0)$ with $\theta(0) \in [-90\degree, 90\degree]$, apply a sequence of joint torques $\u(\cdot)$ such that $\x(T) \approx \x^{des}$, where $T=0.5s$ and $\x^{des}$ is a desired state corresponding to a standing pose with $\theta = 0$.
\end{problem}

\section{Methodology}\label{sec:methodology}

\subsection{Trajectory Optimization}\label{sec:trajectory_optimization}

From inspection of the robot dynamics (\ref{eq:dynamics}), it is clear that the control torques $\btau$ do not have a direct influence on the orientation $\theta$. How then are we to design a control sequence $\btau(\cdot)$ that drives $\theta$ to zero? We have a good understanding of how cats move to achieve such reorientation \cite{kane1969dynamical}, but directly imitating nature is not an option since our robot lacks a flexible spine. Instead, we use trajectory optimization (DDP in particular) to automatically exploit non-holonomic conservation of angular momentum constraints in the dynamics \cite{wieber2006holonomy}. 

\begin{remark}
    We focus on trajectory optimization rather than classical non-holonomic motion planning methods \cite{de1995modelling,murray1993nonholonomic} since such methods typically require the dynamics to be expressible in particular forms. These forms are often restrictive and the algorithms based upon them don't include considerations such as torque or joint angle limits. This makes trajectory optimization a more practical choice.
\end{remark}

DDP (and trajectory optimization in general), finds a trajectory for a dynamical system (\ref{eq:dynamics_general}) that minimizes a given cost:
\begin{align}
    \min_{\u(\cdot)} ~& \int_{t=0}^T l(\x(t),\u(t),t) {\rm d}t \\
    \text{s.t. } & \dot{\x} = f(\x,\u) \\
                 & \x(0) \text{ given},
\end{align}
where $l(\x(t),\u(t),t)$ is a smooth (though not necessarily convex) running cost and the dynamics (\ref{eq:dynamics_general}) are also smooth. DDP is particularly efficient because it takes advantage of the structure of the above problem  \cite{mayne1966second,tassa2012synthesis,li2020hybrid}. Constraints like joint limits can be imposed using barrier \cite{hauser2006barrier} or augmented Lagrangian methods \cite{Howell19}. Importantly, the fact that the solver has access to the dynamics model (\ref{eq:dynamics_general}) means that complex constraints in the dynamics, including conservation of angular momentum, can be automatically handled by the solver. Like any local optimization method, however, DDP is sensitive to a choice of initial guess, especially in terms of solve times.

\subsection{Supervised Learning}\label{sec:supervised_learning}

While trajectory optimization is well-suited to solve Problem~\ref{prob:falling_cat} in simulation, real-time implementation is a challenge. Specifically, the robot must move from detecting a fall to executing a recovery sequence very rapidly, as there is only a fraction of a second before the robot hits the ground. Solve times for trajectory optimization methods like DDP depend heavily on the quality of an initial guess, especially for nonconvex problems like this one. This makes DDP a popular choice for Model Predictive Control (MPC) where a good initialization is readily available \cite{tassa2012synthesis}, but less ideal for problems like this one where an initial high quality solution must be found rapidly. 

We address this issue by combining offline trajectory optimization with supervised machine learning. Specifically, we know that we can use DDP to solve Problem~\ref{prob:falling_cat} from many initial conditions offline. We then use a neural network to ``memorize'' the solutions and interpolate between them. At runtime, all we need to do is to make predictions using the trained network, which is extremely fast.

There are several possibilities for how exactly to formulate this memorization process. The simplest option is to learn a policy that maps states $\x$ to control torques $\u$ directly, i.e., 
\begin{equation}
    \u(t) = \Psi^{policy}_{\W}\Big(\x(t)\Big),
\end{equation}
where $\Psi^{policy}_{\W}$ is a neural network parameterized by weights $\W$. From offline DDP, we obtain a dataset of optimal state-action pairs ($\x^*, \u^*$). This dataset can then be used with standard regression techniques \cite{mohri2018foundations} to determine locally optimal weights $\W$. At runtime, this policy network is evaluated at each timestep to determine the applied joint torques. 

Alternatively, we can learn a mapping from the initial state to an entire trajectory:
\begin{equation}
    \x(\cdot), \u(\cdot) = \Psi^{reflex}_{\W}\Big( \x(0) \Big),
\end{equation}
where $\Psi^{reflex}_{\W}$ is again a neural network parameterized by weights $\W$. In keeping with the feline inspiration of this work, we call this the \textit{reflex} approach, since the network characterizes a kind of automatic response which brings the robot to the desired level state $\x^{des}$. Of course, this generated trajectory may not be dynamically feasible, since the network has no direct knowledge of the robot's dynamics. We address this issue by tracking the reflex using a PD+ controller:
\begin{equation}\label{eq:pd_plus}
    \btau = \btau^{nom} + \K_P(\q - \q^{nom}) + \K_D(\qd - \qd^{nom}),
\end{equation}
where $\btau^{nom}$, $\q^{nom}$, and $\qd^{nom}$ are obtained from the network, and $\K_P$ and $\K_D$ are proportional and derivative gains. 

\begin{remark}
    For either the reflex or the policy approach, the neural network could be used to generate an initial guess for DDP, as explored in \cite{mansard2018using}. This would eliminate the need for a PD+ tracking controller in the reflex case, and could potentially lead to significant performance improvements. 
\end{remark}

The reflex and policy approaches have distinct advantages and disadvantages. The advantage of a policy is that feedback is inherent, and a successful policy implementation could be more robust to mid-fall disturbances. For the reflex approach, once a recovery trajectory is initiated it cannot be changed to adapt to disturbances or changing conditions.

On the other hand, the reflex approach is significantly less data-intensive, as the network only needs to learn a mapping from a relatively restricted set of possible initial states rather than from any state the robot might encounter during a fall. The reflex approach is also more interpretable, as the generated recovery trajectory can be checked for extreme joint torques or self-collisions before execution. Most importantly, however, the reflex approach is much less sensitive to small errors in the network. Under either approach, control actions generated by the network will always have some (potentially small) error relative to the optimal control actions. Under a policy approach these errors can compound as the policy is rolled out, but this does not occur with the reflex approach as no roll-out is needed \cite{mansard2018using}. We find that this leads to significantly more reliable performance for the reflex approach (see Section~\ref{sec:simulation}), and therefore use the reflex approach exclusively in our hardware experiments (Section~\ref{sec:hardware}). 

\section{Simulation Tests}\label{sec:simulation}

\begin{figure}
    \centering
    \begin{subfigure}[t]{0.6\linewidth}
        \includegraphics[width=\linewidth]{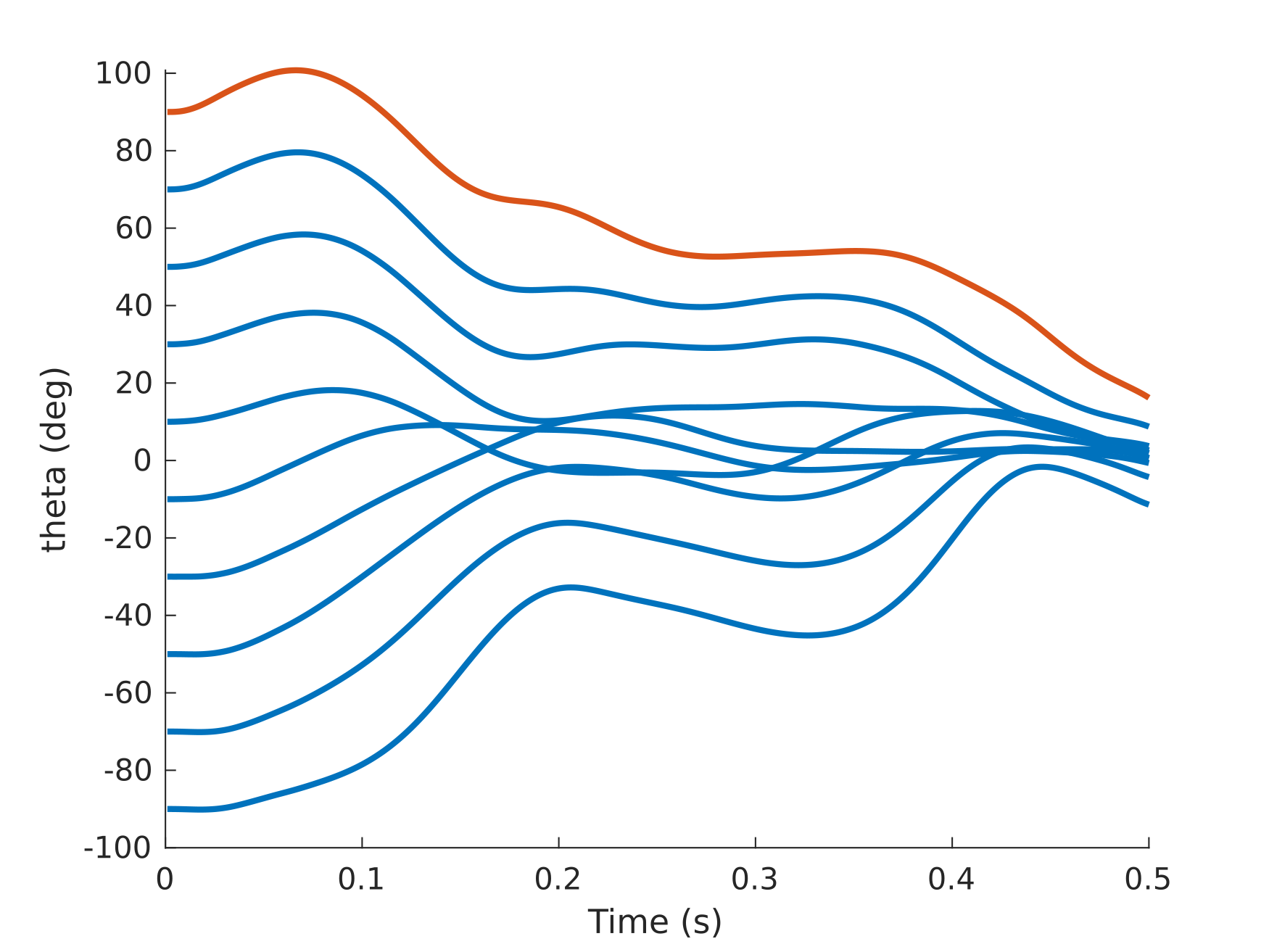}
        \caption{Pitch angle ($\theta$) over time. }
        \label{fig:theta_reflex_plot}
    \end{subfigure}
    \begin{subfigure}[t]{0.35\linewidth}
        \includegraphics[width=\linewidth]{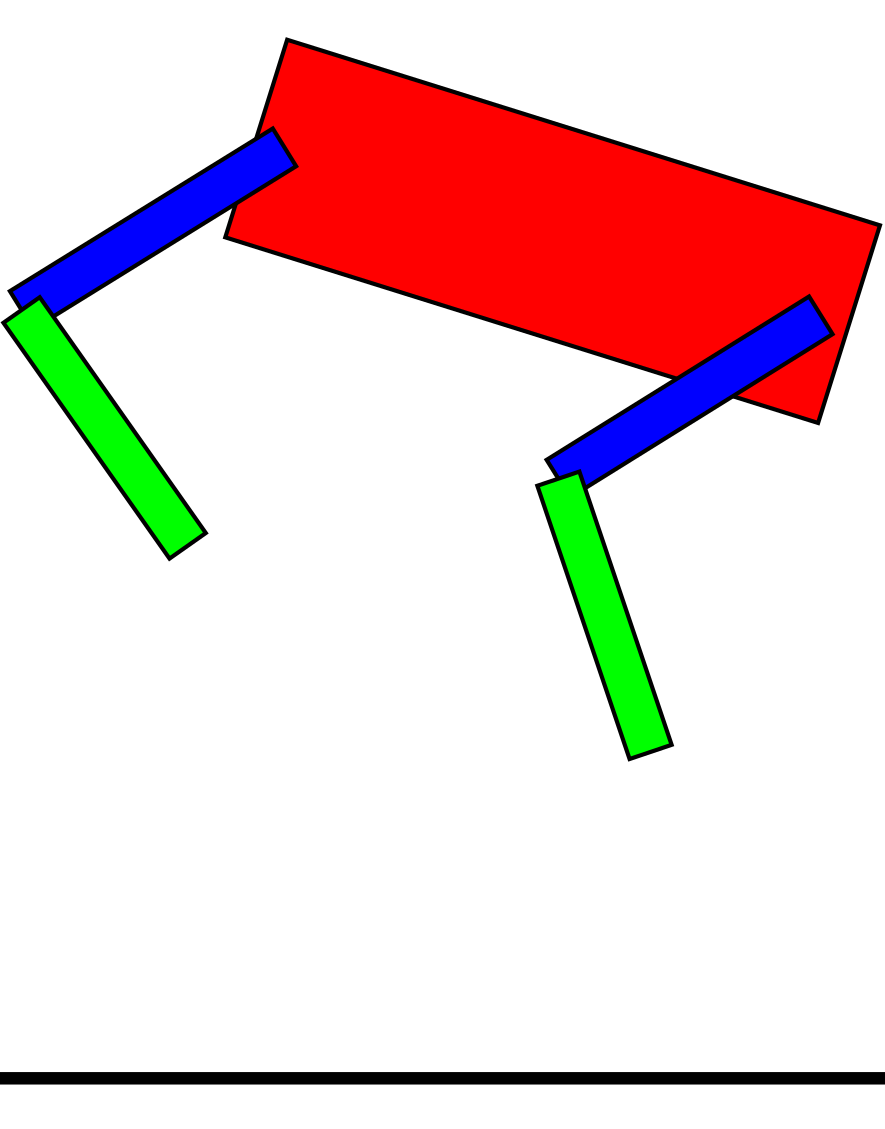}
        \caption{Final configuration from $\theta=90\degree$. }
        \label{fig:theta_reflex_config}
    \end{subfigure}
    \caption{Orientation over time using the ``reflex'' strategy in simulation. A neural network maps an initial orientation to an entire trajectory, which is then tracked with a PD+ controller. The final configuration (\subref{fig:theta_reflex_config}) corresponds to the red highlighted trajectory in (\subref{fig:theta_reflex_plot}).}
    \label{fig:theta_reflex}
\end{figure}

\begin{figure}
    \centering
    \begin{subfigure}[t]{0.6\linewidth}
        \includegraphics[width=\linewidth]{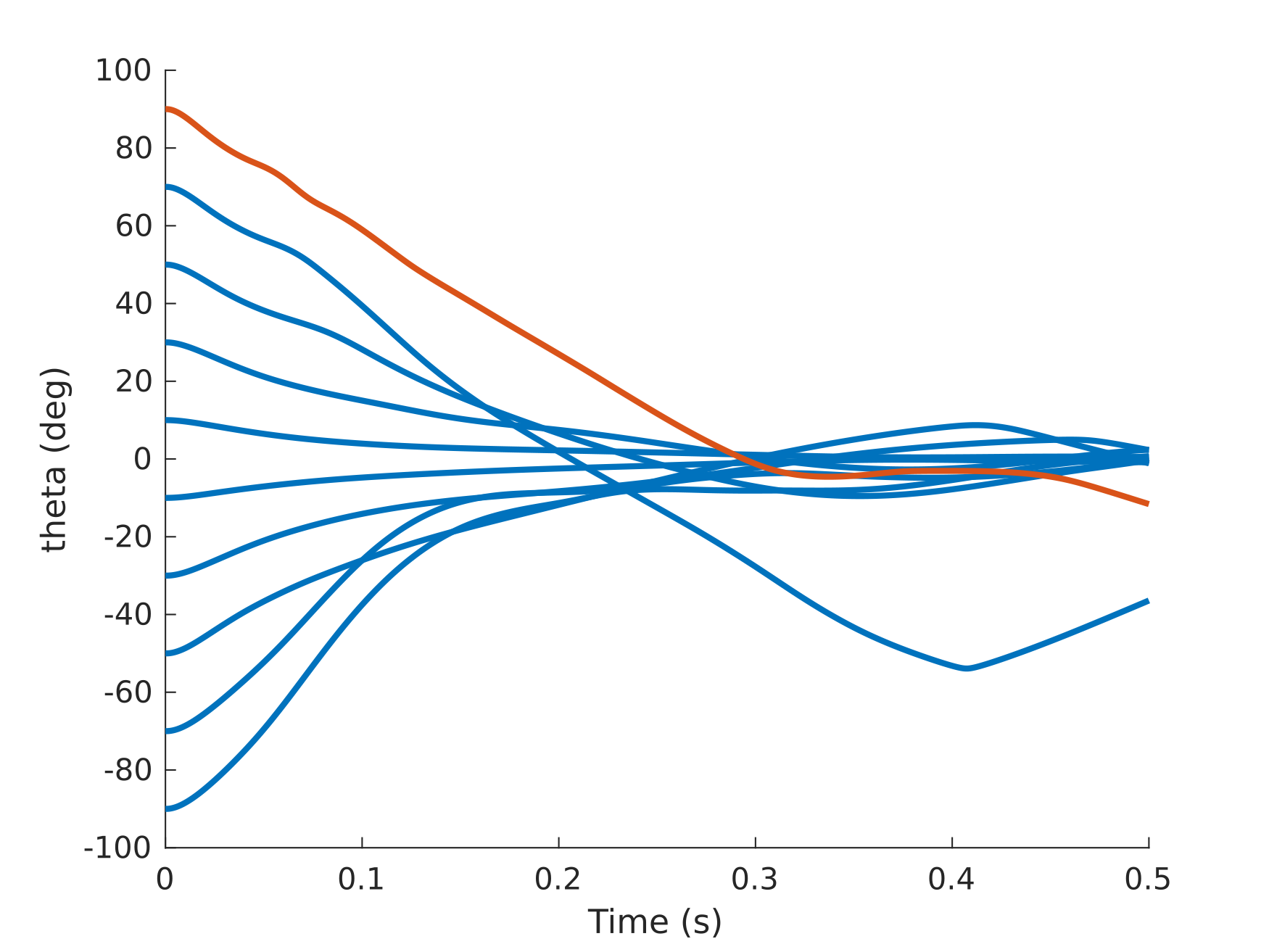}
        \caption{Pitch angle ($\theta$) over time. }
        \label{fig:theta_policy_plot}
    \end{subfigure}
    \begin{subfigure}[t]{0.35\linewidth}
        \includegraphics[width=0.9\linewidth]{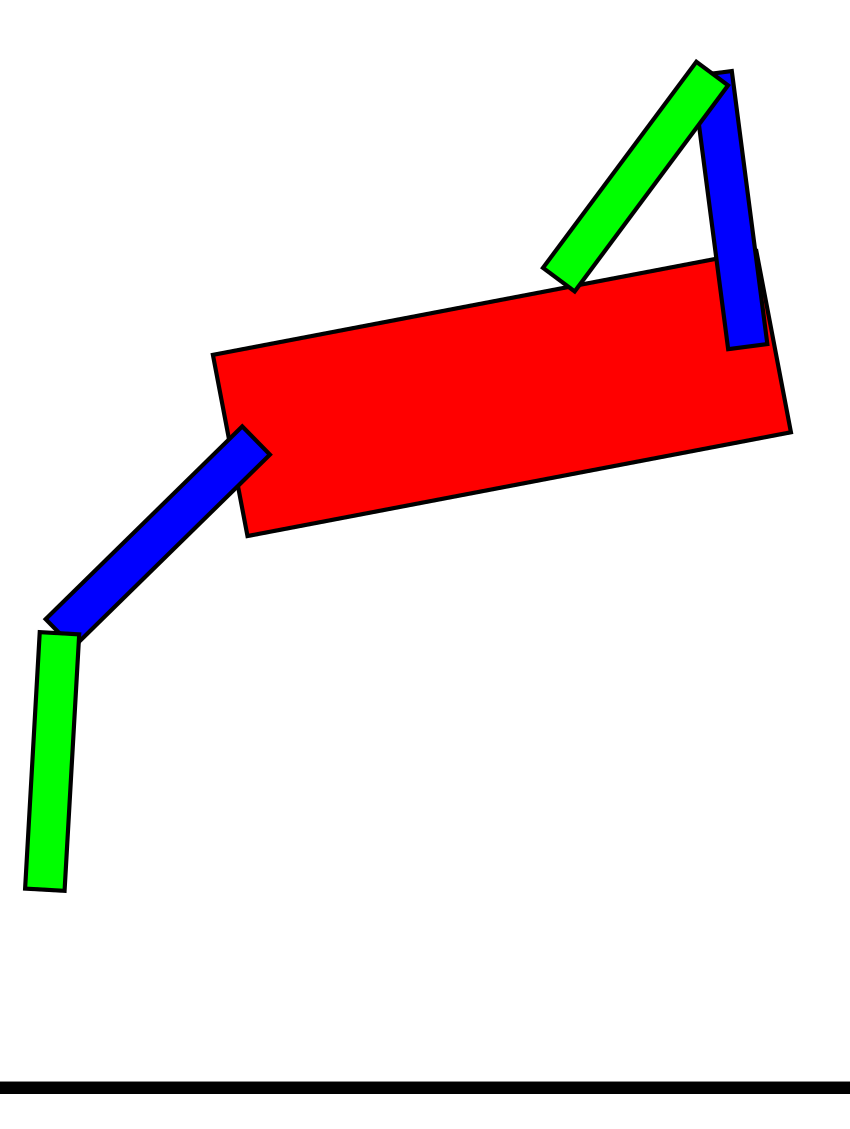}
        \caption{Final configuration from $\theta=90\degree$. }
        \label{fig:theta_policy_config}
    \end{subfigure}
    \caption{Orientation over time using the ``policy'' strategy in simulation. A neural network maps states to control actions at each timestep. The final configuration (\subref{fig:theta_policy_config}) corresponds to the red highlighted trajectory in (\subref{fig:theta_policy_plot}).}
    \label{fig:theta_policy}
\end{figure}

We performed simulation tests using a planar model of the Mini Cheetah (with 500g boots) in MATLAB. This model is depicted in Figures \ref{fig:theta_reflex_config} and \ref{fig:theta_policy_config}. We used the Spatial-v2 \cite{spatialv2} package for dynamics computations and a MATLAB implementation of \cite{li2020hybrid} for trajectory optimization. \hl{4th-order Runge-Kutta integration was used with a 1ms timestep.}

For trajectory optimization, we used the cost function
\begin{equation}\label{eq:cost}
    \int_{t=0}^{T-1} \bigg( \|\x(t) - \x^{des}\|^2_{\Q} +  \|\u(t)\|^2_{\R} \bigg){\rm d}t + \|\x(T) - \x^{des}\|^2_{\Q_f} 
\end{equation}
with weights $\Q$ and $\Q_f$ designed to prioritize orientation during the fall and joint angles at the end of the fall. Fairly restrictive joint limits were chosen such that no self-collisions were possible, and were imposed using a relaxed barrier strategy \cite{hauser2006barrier, li2020hybrid}. Solving the trajectory optimization problem from scratch typically took between 60 and 200 iterations, corresponding to a wall-clock time of 45-150 seconds\footnote{Note that solve times could be significantly improved with a more efficient C/C++ implementation.}.

For both the reflex and policy networks, we used simple feed-forward networks with two hidden layers, each with 512 units and ReLU activation functions. We implemented these networks in Tensorflow \cite{tensorflow2015-whitepaper} and used the Adam optimizer \cite{kingma2014adam} with default options. 

For the reflex network, we generated a dataset of consisting of 400 pairs of initial conditions and subsequent optimal trajectories. The initial conditions consisted of randomly sampled orientations in $[-90\degree, 90\degree]$ and joint angles corresponding to a standing position. Each trajectory consisted of the robot state $\x$ and input $\u$ at 1ms intervals for 0.5s. When generating the dataset, we sorted the initial conditions and used prior optimal trajectories to warm-start the DDP solver. This reduced DDP solve times considerably, down to 1-2 seconds per trajectory. 

To train the policy network, we converted this dataset into optimal state-action pairs (500 pairs per trajectory, 400 trajectories). The resulting policy network performed extremely poorly, generating only seemingly random flailing motions, so we augmented the policy dataset with 600 additional trajectories. These additional trajectories were initialized with random joint angles, joint velocities, and angular velocities. 

Results of applying the reflex strategy in simulation from a variety of initial orientations are shown in Fig.~\ref{fig:theta_reflex}. Pitch angles converge consistently to approximately level, \hl{with a mean final pitch of $-0.9\degree$ and standard deviation of $9.7\degree$}. The final configuration of the robot for a representative trajectory (highlighted in red) is shown in (\ref{fig:theta_reflex_config}). Note that the legs are positioned under the robot, ready for a soft landing. 

Similar results for the policy strategy are shown in Fig.~\ref{fig:theta_policy}. Orientations generally converge faster than under the reflex strategy, except for one of the trajectories, which performs very badly. Even for the trajectories that seem to perform well, however, the final joint angles are sometimes far from the nominal standing position, as shown in (\ref{fig:theta_policy_config}). We suspect that this behavior is due to the compounding of small inaccuracies in the policy network when the policy is rolled out. Similar issues with a policy-based approach are discussed in \cite{mansard2018using}. 

\begin{figure*}
    \centering
    \includegraphics[width=0.15\linewidth]{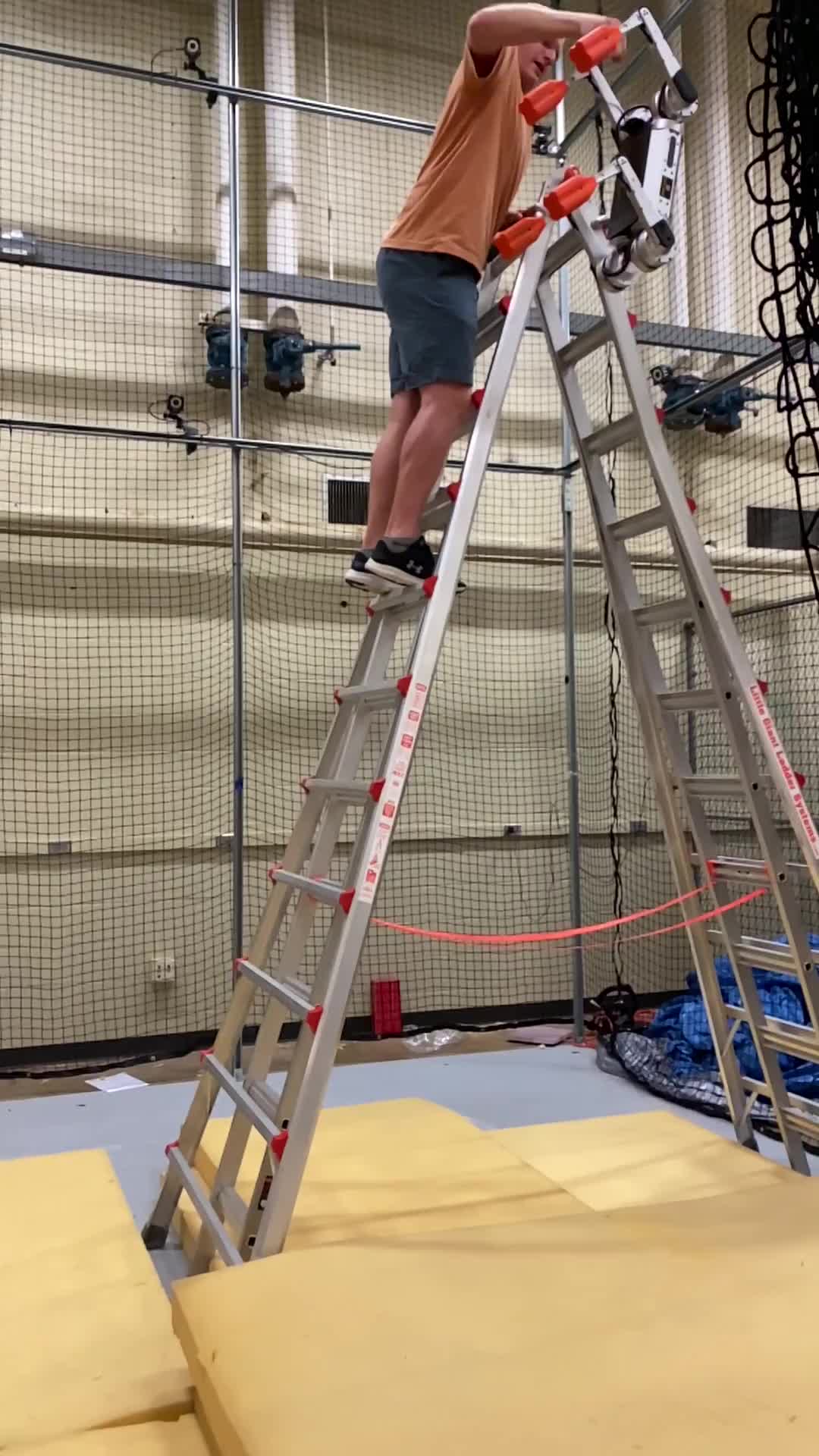}
    \includegraphics[width=0.15\linewidth]{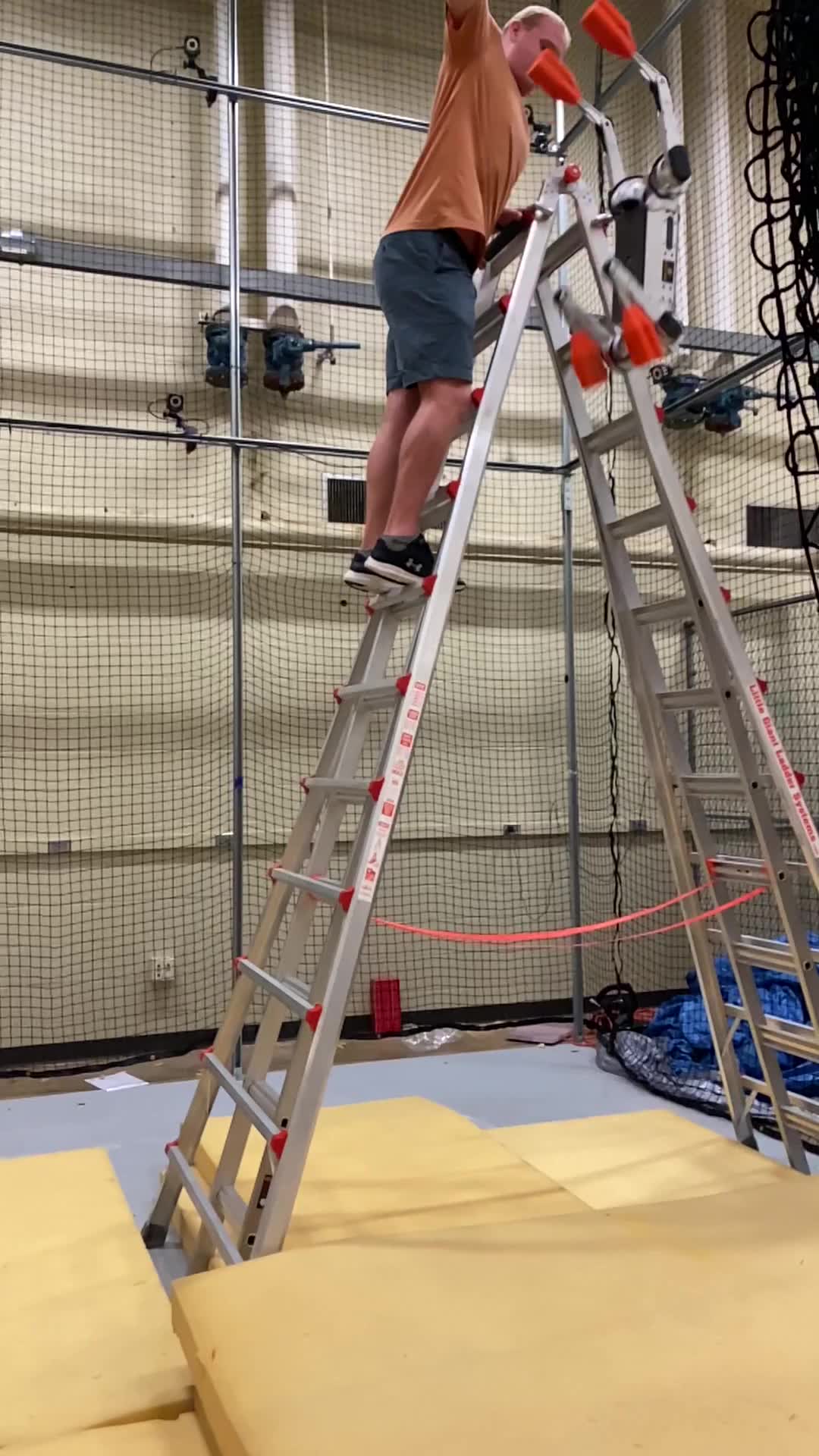}
    \includegraphics[width=0.15\linewidth]{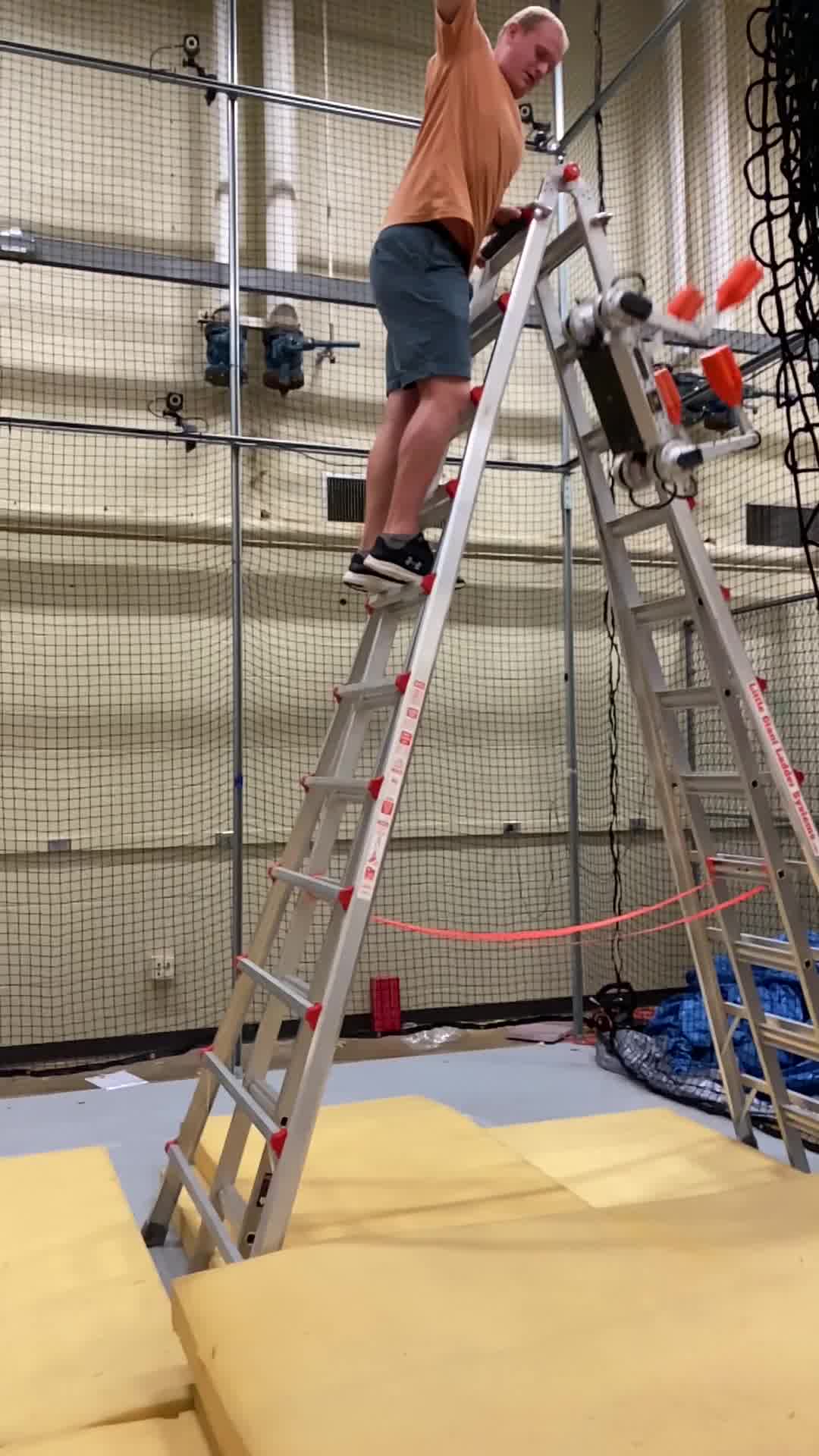}
    \includegraphics[width=0.15\linewidth]{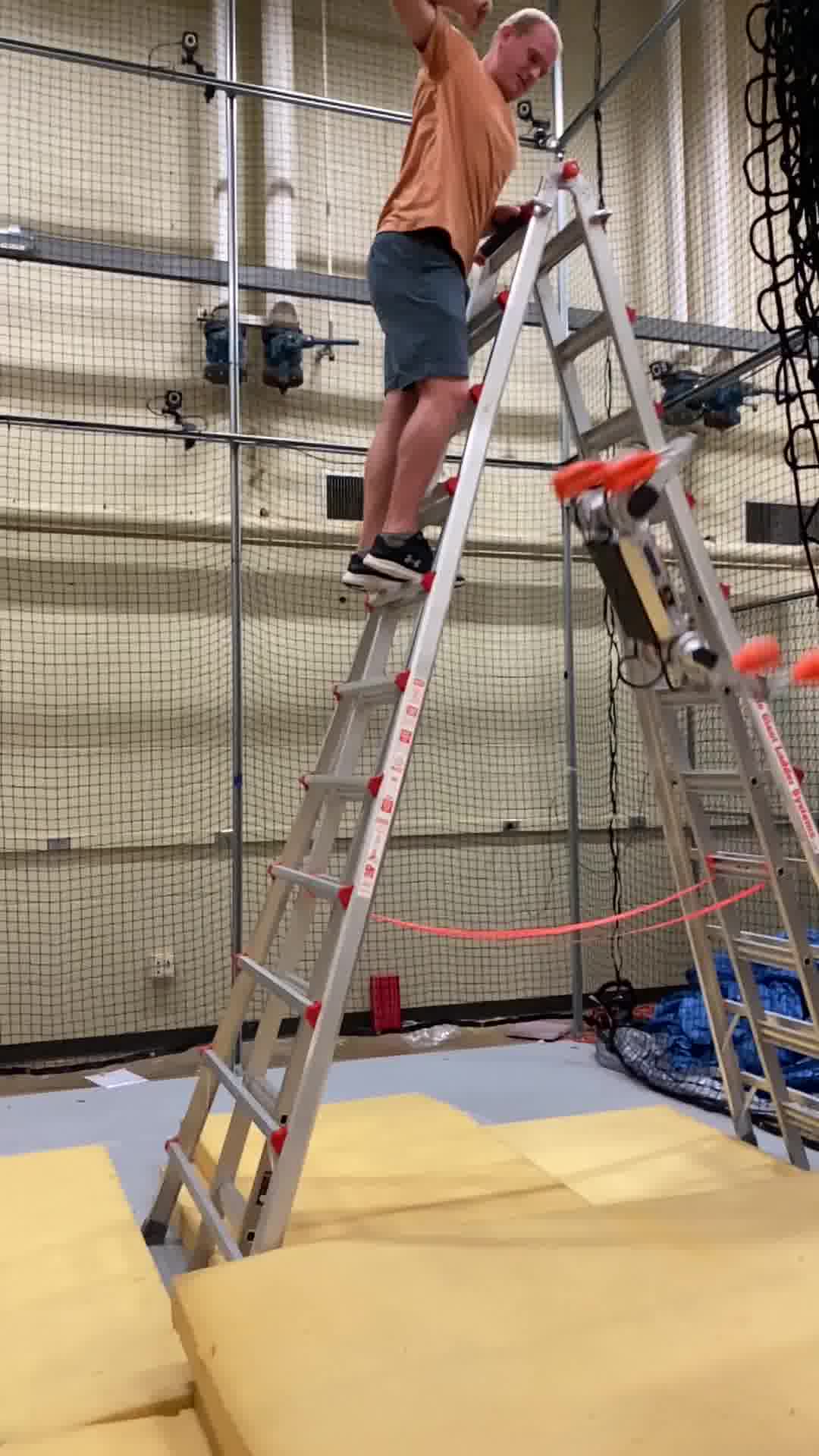}
    \includegraphics[width=0.15\linewidth]{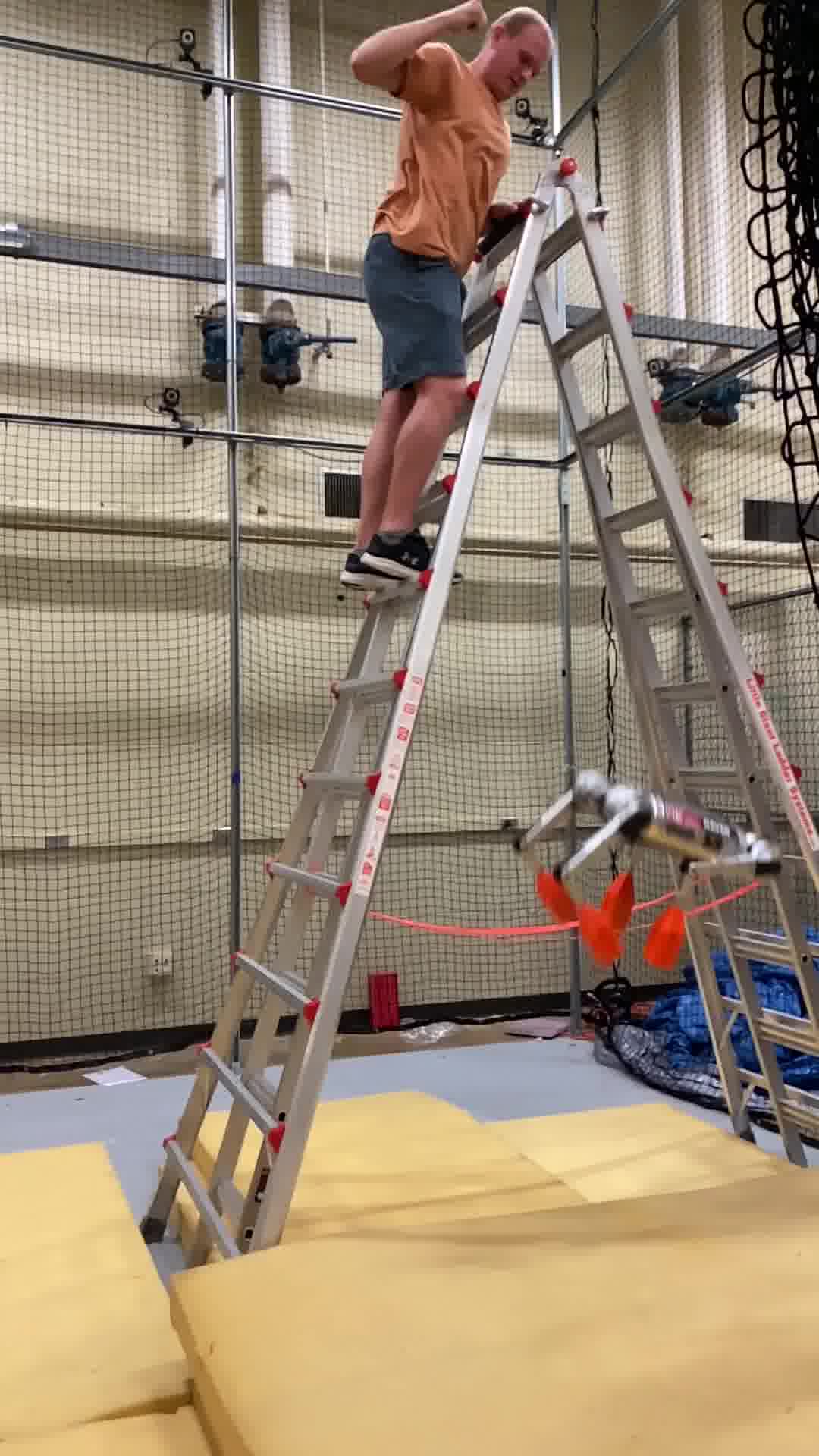}
    \includegraphics[width=0.15\linewidth]{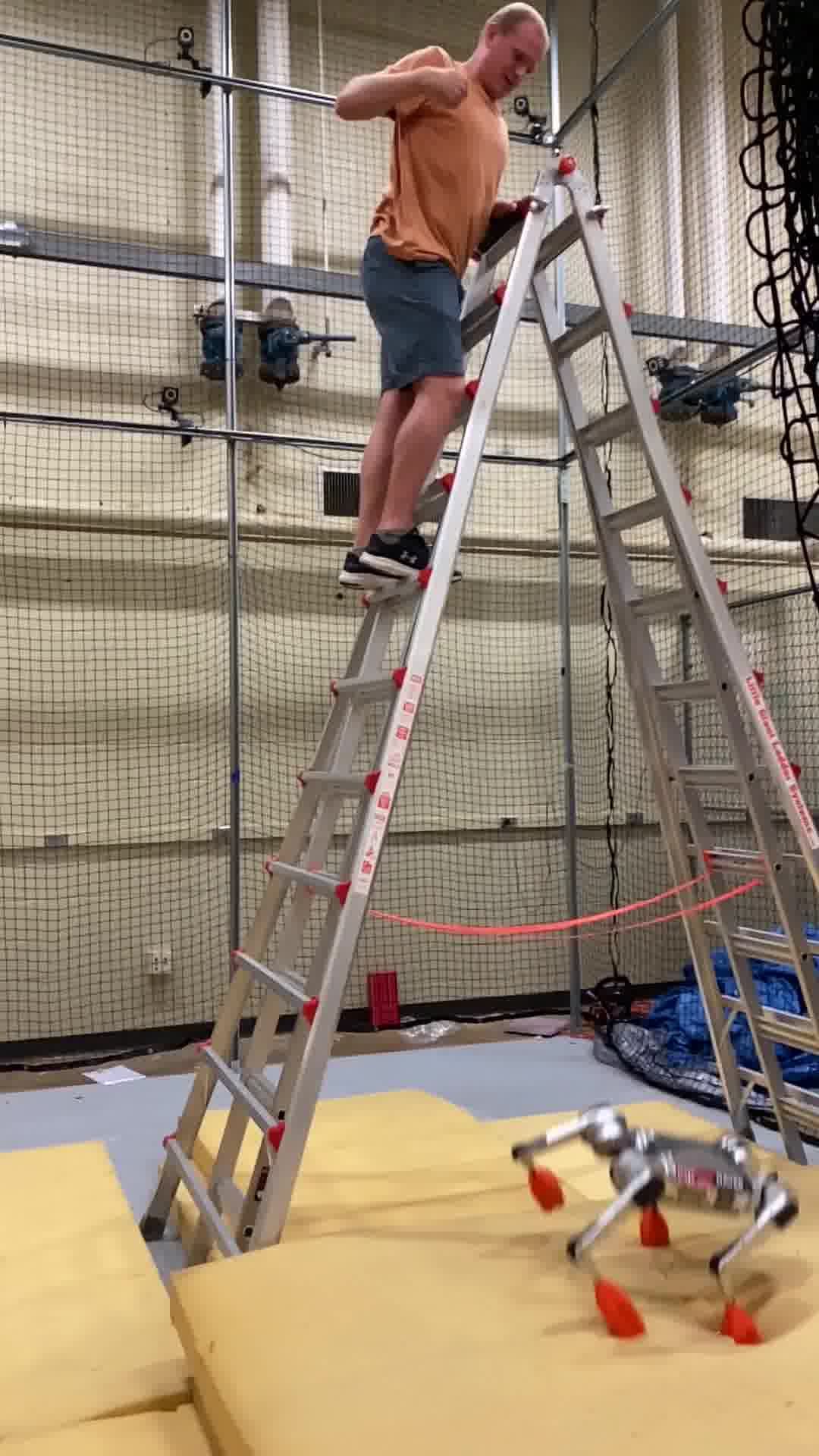}
    \caption{Video frames from one of the hardware tests. The robot was dropped from various orientations onto a stack of soft mats.}
    \label{fig:hardware_setup}
\end{figure*}

Another possible explanation for the poor performance of the policy method, documented here and in \cite{mansard2018using}, comes from the fact that non-holonomic systems cannot be stabilized with smooth time-invariant control laws \cite{brockett1983asymptotic,de1995modelling,wieber2006holonomy}. Our policy network is essentially a time-invariant control law, though not necessarily a smooth one (as long as non-smooth activation functions are used in the policy network). 


While the problems we found with the policy-based approach could likely be mitigated with additional training data and hyperparameter tuning, these challenges---especially regarding violations of joint limits---motivate us to focus exclusively on the reflex strategy for hardware implementation. 

\section{Hardware Experiments}\label{sec:hardware}

To implement our reflex-based falling controller on the robot, we first converted the Tensorflow neural network to run in C++ using the frugally-deep software package \cite{hermann2016frugally}. After this conversion, it takes only 2ms to generate a recovery trajectory from a given initial condition. We then added a simple ``falling cat'' mode to Mini Cheetah's existing control architecture \cite{katz2019mini}. 

This falling cat mode starts by applying a simple joint-space PD controller that holds the legs in a nominal standing position. Once we are ready to drop the robot, a switch on the remote controller indicates to the robot that it should start trying to detect a fall. A fall is detected once the IMU indicates base accelerations of $9.81 \pm 0.1 m/s^2$ for 15 timesteps in a row (30ms). Before a fall is detected, the robot continues to apply the PD control law. As soon as a fall is detected, the robot uses the reflex network to generate a $0.5s$ recovery trajectory from its current state estimate. This recovery trajectory is then tracked using (\ref{eq:pd_plus}). Once the recovery trajectory is finished, we apply Mini Cheetah's standard balancing controller with additional PD compensation. This force-based controller attempts to balance the robot on its feet: the additional PD compensation ensures that the motion of the legs remains reasonable even if the robot hasn't reached the ground yet. 

\begin{figure}
    \centering
    \includegraphics[width=0.8\linewidth]{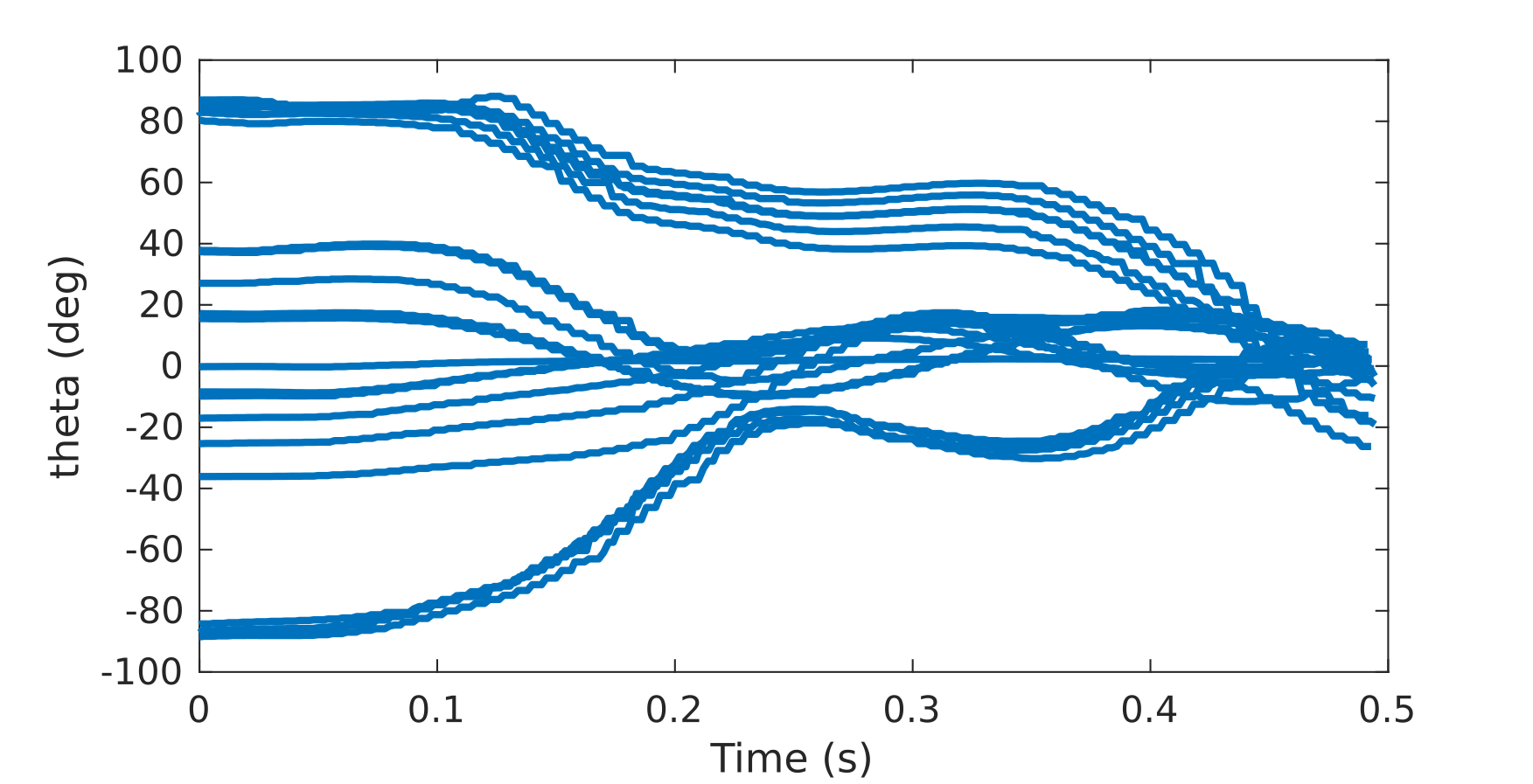}
    \caption{Pitch angle ($\theta$) over time for hardware experiments.}
    \label{fig:theta_hardware}
\end{figure}

We dropped the robot from various initial orientations onto a stack of soft mats from a height of about 10ft, as shown in Fig.~\ref{fig:hardware_setup}. \hl{At this height, the 0.5s recovery trajectory is completed well before the robot hits the ground}. A plot of pitch angle over time is shown in Fig.~\ref{fig:theta_hardware}. This matches the simulation fairly well, \hl{with mean final pitch of $-2.2\degree$ and standard deviation $10.6\degree$}. We believe that most of this additional error is due to modeling inaccuracy, especially from the heavy boots, and the fact that the robot does not actually operate in a plane. 

\begin{figure}
    \centering
    \begin{subfigure}{0.48\linewidth}
        \includegraphics[width=\linewidth]{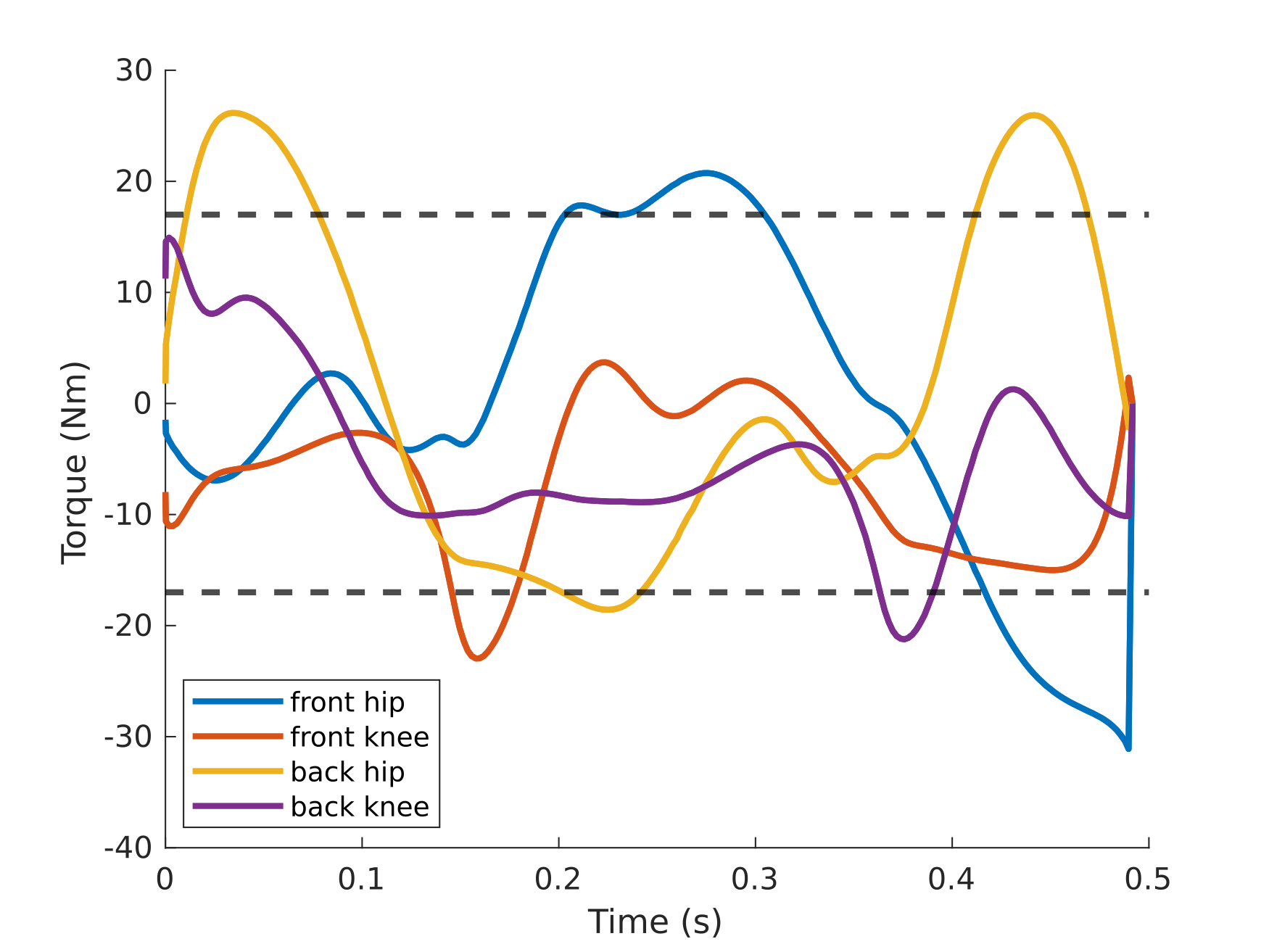}
        \caption{Network-generated torques $\btau^{nom}$}
        \label{fig:tau_ff} 
    \end{subfigure}
    \begin{subfigure}{0.48\linewidth}
        \includegraphics[width=\linewidth]{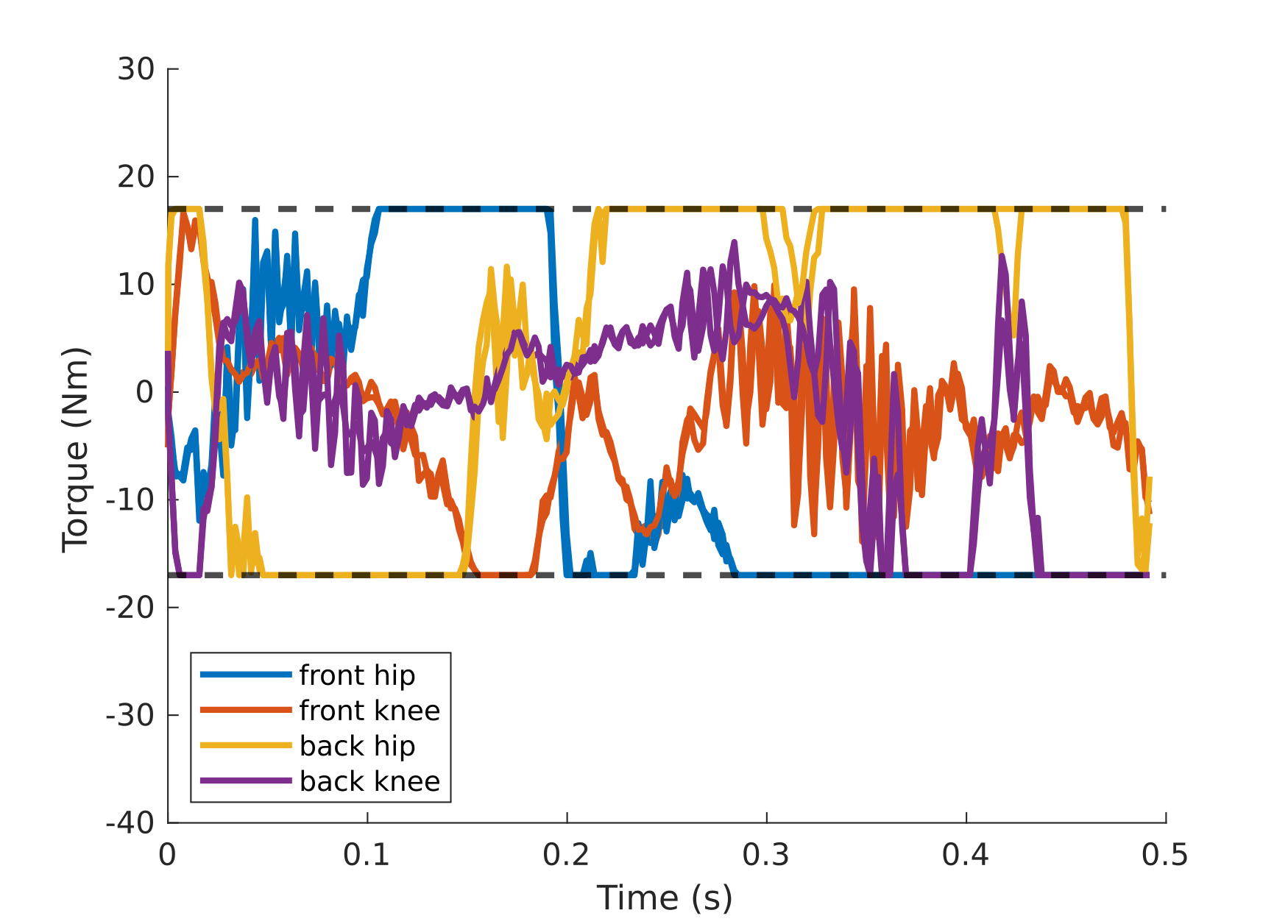}
        \caption{Applied torques $\btau$ (\ref{eq:pd_plus})}
        \label{fig:tau_actual} 
    \end{subfigure}
    \caption{Joint torques generated by the reflex network (\subref{fig:tau_ff}) and applied on the robot (\subref{fig:tau_actual}) during a fall from roughly $-90\degree$ (facing directly down). Black dotted lines indicate torque limits. Differences between $\btau$ and $\btau^{nom}$ come from the fact that the network-generated trajectory is not dynamically feasible and from modeling error, especially with regard to the heavy boots.}
    \label{fig:torque_profile}
\end{figure}

\begin{figure}
    \centering
    \includegraphics[width=0.9\linewidth]{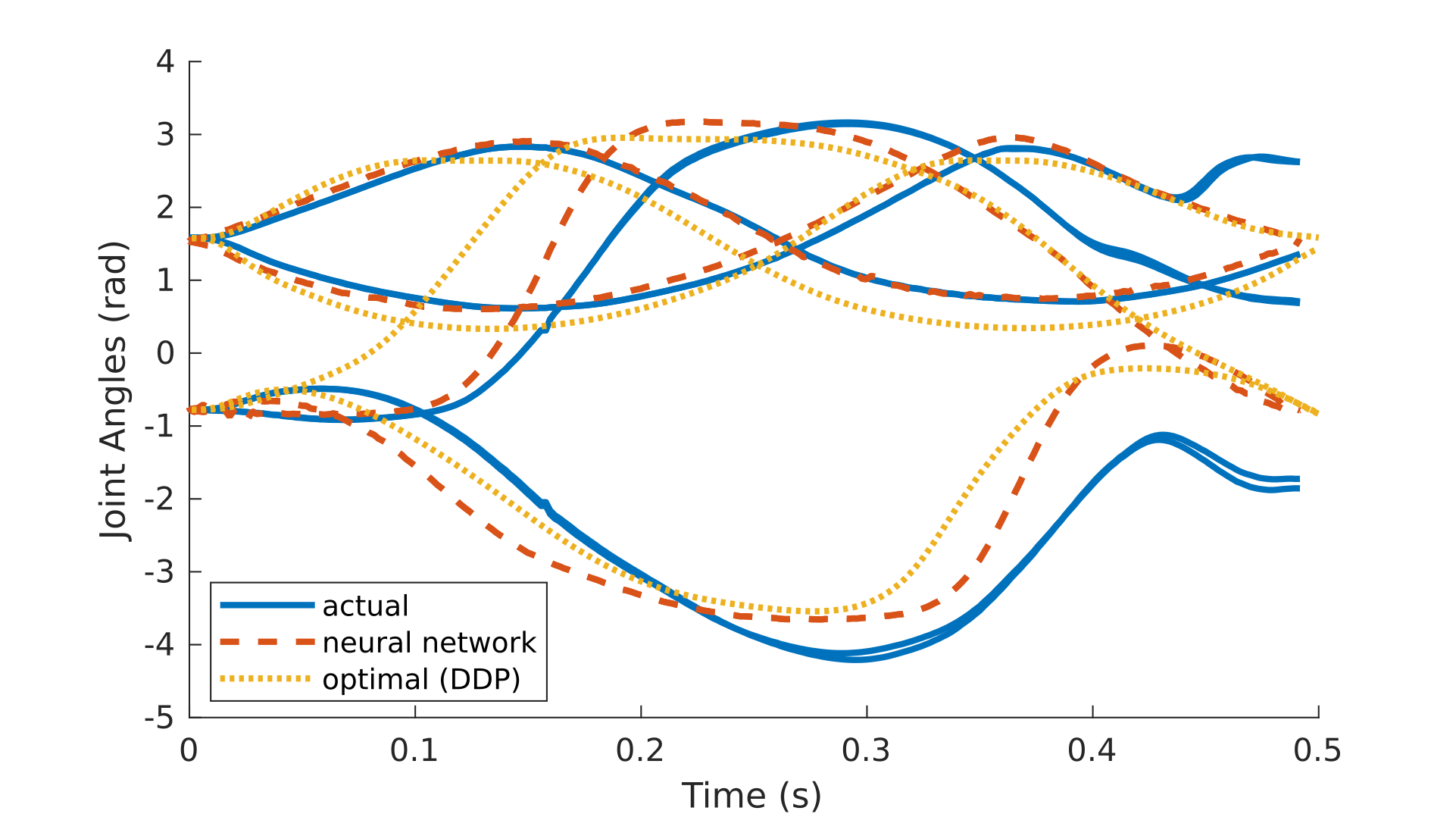}
    \caption{Actual joint angles ($\q$), desired joint angles generated by the reflex network ($\q^{nom}$), and optimal joint angles from trajectory optimization ($\q^{opt}$) over time for a fall from roughly $-90\degree$ (facing directly downwards).}
    \label{fig:joint_profile}
\end{figure}

Applied joint torques $\btau$ over time for one of the experiments are shown in Fig.~\ref{fig:torque_profile}. Actual joint angles $\q$, desired joint angles $\q^{nom}$ generated by the reflex network, and optimal joint angles $\q^{opt}$ generated by solving DDP for the same trajectory are shown in Fig.~\ref{fig:joint_profile}. The primary discrepancy between $\q$ and $\q^{nom}$ is likely due to modeling error. Specifically, our simulation model assumes the heavy boots are 500g point masses located at the feet, when in reality the boots have non-negligible rotational inertia, weigh 532g, and extend beyond the usual position of the feet. The discrepancy between $\q^{nom}$ and $\q^{opt}$ could likely be reduced with additional training data and hyperparameter tuning.

\begin{figure}
    \centering
    \includegraphics[width=\linewidth]{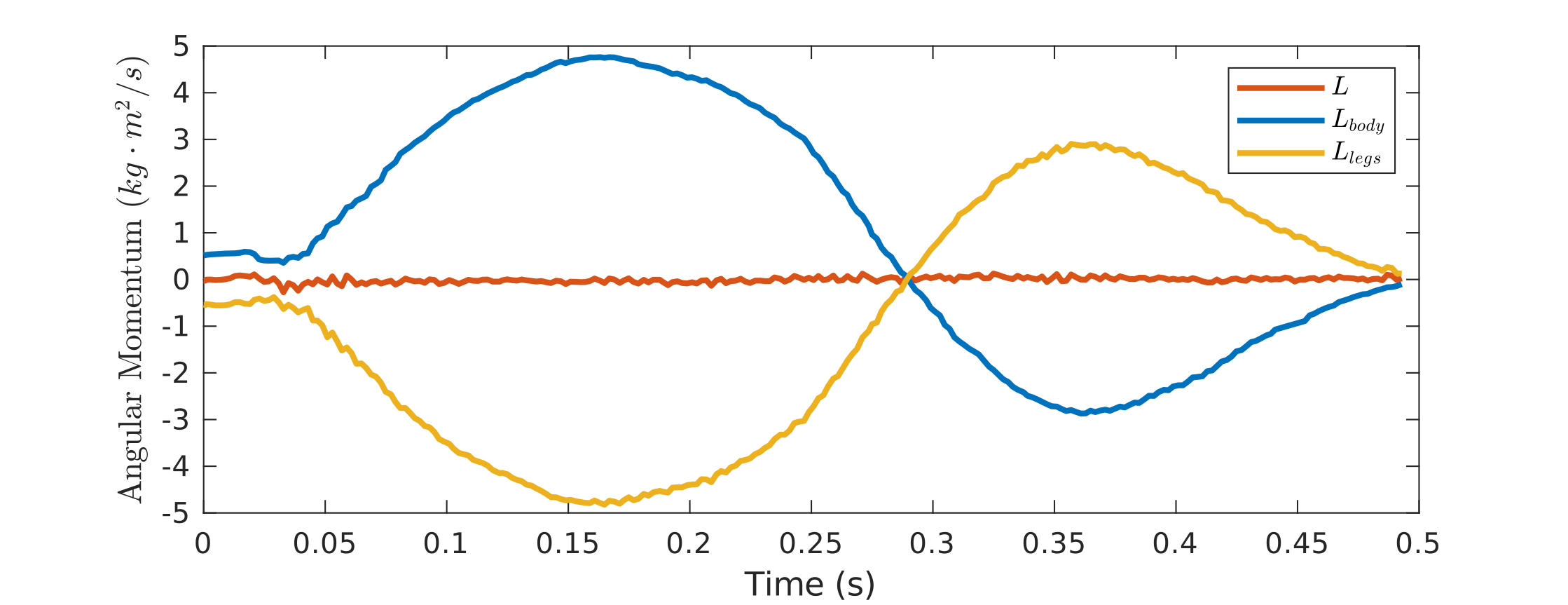}
    \caption{The angular momentum contribution of the legs (yellow) counteracts the angular momentum contribution of the body (blue), allowing the robot to rotate in midair while net angular momentum (red) is conserved. }
    \label{fig:am}
\end{figure}

Fig.~\ref{fig:am} depicts the net angular momentum about the pitch axis ($L$) together with the contributions of the body ($L_{body}$) and the legs ($L_{legs}$). This demonstrates that $L_{body}$ is modulated by $L_{legs}$ while $L$ is conserved.

\section{Discussion}\label{sec:discussion}

In the previous sections, we showed how a combination of trajectory optimization and machine learning can be used to land a quadruped robot on its feet like a cat. In this section we discuss some of the limitations of our control approach, and potential areas of future research to address these limitations. 

One notable limitation is that our robot only lands on its feet from an initial pitch angle between $-90\degree$ and $90\degree$, whereas cats can land on their feet when dropped from fully upside down. Additionally, cats need only a few feet of freefall to land on their feet, whereas our robot was dropped from a ladder 10ft above the ground.

One reason for this difference may be the limitations of the hardware platform. The Mini Cheetah robot is considerably heavier than a cat, especially with added boots, and required operation at the torque limits to perform $90\degree$ of rotation (Fig.~\ref{fig:torque_profile}). Perhaps more critical, however, was our choice to consider rotation only in the sagittal plane. It is reasonable to expect that better performance might be obtained by rotating side-to-side, where the inertia of the body is lower. 

We chose to consider only planar motion because of the challenge of self-collisions. Self-collisions are highly non-convex and difficult to incorporate efficiently in a trajectory optimization framework. In this work we avoided self-collisions by imposing (much simpler) joint angle limits. This strategy is very conservative, however, and would likely eliminate many useful motions if the full 3D dynamics of the robot were considered. To this end, efficient methods of encoding such highly nonconvex constraints in a trajectory optimization framework are an important area of future research. 

Another limitation of our proposed approach stems from the well-documented robustness issues when it comes to neural networks \cite{szegedy2013intriguing}. Small perturbations to the network's input can result in drastically different outputs. This has the potential to cause considerable harm in our case, since the network output is a high-torque trajectory that is executed on the robot. The interpretability of the reflex strategy mitigates this risk somewhat, as the trajectory can be checked for self-collisions or extreme joint torques before application on the robot. Recent results on verifying neural network robustness using Satisfiability Modulo Theories \cite{katz2017reluplex} or Mixed Integer Programming \cite{tjeng2018evaluating} could also be leveraged to reduce this risk. 

A more satisfying approach might be to use the network output as an initial guess for a trajectory optimizer. This would help address the adversarial robustness issue, as the final trajectory would be a locally optimal one generated by DDP. Furthermore, this could lead naturally to applying trajectory optimization in an MPC fashion during the fall.  

A larger research question is how to best store information from offline trajectory optimization. This work suggest that a reflex-based approach is superior to a policy-based approach, at least for short and highly dynamic trajectories, but these are not the only possible choices. One might also learn an approximation of the optimal value function, or use some combination of policy-based and reflex-based approaches. This is closely related to the choice of action parameterization in RL. Interestingly, a popular action parameterization in RL for legged locomtion is joint angles for several timesteps into the future \cite{peng2020learning}. This is in some sense similar to a trajectory, suggesting that there may be something about state projections into the future that is ideal for storing learned motions.

Finally, a major challenge in our hardware implementation was the landing itself. Our trajectory design process considered only reorientation in flight, not the considerably more complex contact phase with the ground. In practice, this often led to the robot striking the mat feet-first, only to bounce up and land on its back. This could be mitigated by using contact-aware trajectory optimization strategies like \cite{li2020hybrid,patel2019contact,sleiman2019contact,manchester2020variational}, particularly if they are applied in MPC fashion. A related challenge is contact detection, as many contact detection algorithms \cite{bledt2018contact} are tailored for rigid contacts. 

\section{Conclusion}\label{sec:conclusion}

We demonstrated how a combination of trajectory optimization and supervised machine learning can enable a quadruped robot to land on its feet like a cat. We considered two potential schemes for learning offline-generated optimal motions and found that a reflex-based strategy was more reliable than a policy-based strategy. While the landing task required operation at the torque limits of the Mini Cheetah, we believe that algorithmic challenges rather than hardware limitations are the primary limiting factor in achieving even more extreme robot acrobatics. Specifically, further research in trajectory optimization with self-collision constraints, safe machine learning, and contact-aware control is needed to match the astounding abilities of cats. 

\section{Acknowledgements}

Thanks to Angela Rauch and Robert Frei for their help in designing the 3D printed boots, and Shenggao Li for assisting with the hardware. The Mini Cheetah is sponsored by the MIT Biomimetic Robotics Lab and NAVER LABS.

\newpage

\bibliographystyle{IEEEtran}
\balance
\bibliography{references}

\newpage
\nobalance

\section*{Appendix A\\Cost Function Details}

The cost function (\ref{eq:cost}) was specified using the following parameters:
\begin{equation*}
    \x^{des} = 
    [\begin{array}{c;{2pt/2pt}c;{2pt/2pt}cccc;{2pt/2pt}c}
        \undermat{\theta}{~~0~} & \undermat{\dot{\theta}}{~~0~} & \undermat{\q}{0.778 & -1.578 & 0.778 & -1.578} & \undermat{\dot{\q}}{~~\bm{0}~}
    \end{array}]
\end{equation*}
~
\begin{align*}
    \R &= 1.0\bm{I}_{4\times4}, \\
    \Q &= \text{diag}([
    \begin{array}{c;{2pt/2pt}c;{2pt/2pt}c;{2pt/2pt}cccc}
        2000 & 0.1 & 20 & 0.1 & 10 & 0.1 & 10
    \end{array}]), \\
    \Q_f &= \text{diag}([
    \begin{array}{c;{2pt/2pt}c;{2pt/2pt}c;{2pt/2pt}cccc}
        7000 & 0.1 & 3000 & 0.1 & 0.1 & 0.1 & 0.1
    \end{array}]).
\end{align*}

Note that the largest terms relate to the body orientation $\theta$, especially at the end of the trajectory, and the joint angles at the end of the trajectory. 

\section*{Appendix B\\Neural Network Hyperparameters}

Hyperparameters used to train the reflex approach are shown in Table~\ref{tab:reflex_hyperparams} below:

\begin{table}[h]
    \centering
    \begin{tabular}{c|c}
        Hyperparameter & Value  \\
        \hline
        Epochs & 1000 \\
        Batch size & 2 \\
        Input size & 10 \\
        Output size & 7000 \\
        Learning rate & 0.001 \\
        Hidden layers & 2 $\times$ 512 hidden units \\
        Activation function & ReLU (linear output) \\
        Optimizer & Adam
    \end{tabular}
    \caption{Reflex Network Hyperparameters}
    \label{tab:reflex_hyperparams}
\end{table}

All other parameters were Tensorflow defaults. 
A dataset of 400 trajectories was used to train the network, with 10\% (40 trajectories) held back as a validation dataset. Initial orientations were uniformly sampled between $-90$ and $90$ degrees. Training the reflex network took about 15 minutes on a workstation laptop with an Intel i7 processor, Nvidia M1000M GPU, and 32 GB RAM. 

Hyperparameters for the policy approach are shown in Table~\ref{tab:policy_hyperparams} below:

\begin{table}[h]
    \centering
    \begin{tabular}{c|c}
        Hyperparameter & Value  \\
        \hline
        Epochs & 400 \\
        Batch size & 500 \\
        Input size & 10 \\
        Output size & 4 \\
        Learning rate & 0.001 \\
        Hidden layers & 2 $\times$ 512 hidden units \\
        Activation function & ReLU (linear output) \\
        Optimizer & Adam
    \end{tabular}
    \caption{Policy Network Hyperparameters}
    \label{tab:policy_hyperparams}
\end{table}

A dataset of 1000 trajectories was used to train the policy network, again with a 10\% validation dataset. Training the policy network took about 60 minutes on the same laptop.

We did not perform extensive or systematic hyperparameter tuning for either approach. As with any supervised learning application, it is possible that better hyperparameter choices could dramatically improve performance for either the policy or the reflex approach.

\end{document}